\documentclass[sigconf, screen]{acmart}
\usepackage{subfigure}
\usepackage{arydshln}
\usepackage{listings}
\usepackage{threeparttable}
\usepackage{hyperref}
\usepackage{multirow}
\usepackage{xcolor}

\AtBeginDocument{%
  }

\setcopyright{acmlicensed}
\copyrightyear{2026}
\acmYear{2026}
\acmDOI{XXXXXXX.XXXXXXX}
\acmBooktitle{Companion Proceedings of the 34th ACM Symposium on the Foundations of Software Engineering (FSE '26), June 5--9, 2026, Montreal, Canada}
\acmISBN{978-1-4503-XXXX-X/2018/06}

\begin{document}

\title{Help Without Being Asked: A Deployed Proactive Agent System for On-Call Support with Continuous Self-Improvement}

\author{Fengrui Liu}
\email{liufengrui.work@bytedance.com}
\affiliation{%
  \institution{ByteDance}
  \city{Beijing}
  \country{China}
}

\author{Xiao He}
\email{xiao.hx@bytedance.com}
\affiliation{%
  \institution{ByteDance}
  \city{Hangzhou}
  \country{China}
}

\author{Tieying Zhang}
\email{tieying.zhang@bytedance.com}
\affiliation{%
  \institution{ByteDance}
  \city{San Jose}
  \country{United States}
}
\authornote{Corresponding author.}

\renewcommand{\shortauthors}{Fengrui Liu et al.}
\newcommand{\name}{\textit{Vigil}}
\begin{abstract}

In large-scale cloud service platforms, thousands of customer tickets are generated daily and are typically handled through on-call dialogues.
This high volume of on-call interactions imposes a substantial workload on human support analysts.
Recent studies have explored reactive agents that leverage large language models as a first line of support to interact with customers directly and resolve issues.
However, when issues remain unresolved and are escalated to human support, these agents are typically disengaged.
As a result, they cannot assist with follow-up inquiries, track resolution progress, or learn from the cases they fail to address.

In this paper, we introduce \name{}, a novel proactive agent system designed to operate throughout the entire on-call life-cycle.
Unlike reactive agents, \name{} focuses on providing assistance during the phase in which human support is already involved.
It integrates into the dialogue between the customer and the analyst, proactively offering assistance without explicit user invocation.
Moreover, \name{} incorporates a continuous self-improvement mechanism that extracts knowledge from human-resolved cases to autonomously update its capabilities.
\name{} has been deployed on Volcano Engine, ByteDance’s cloud platform, for over ten months, and comprehensive evaluations based on this deployment demonstrate its effectiveness and practicality.
The open source version of this work is publicly available at \url{https://github.com/volcengine/veaiops}.
\end{abstract}




\maketitle

\section{Introduction}

\begin{figure*}[t]
    \centering
    \subfigure[Reactive Agent: Responds to every query directly until escalation.]{
        \includegraphics[width=.4\textwidth]{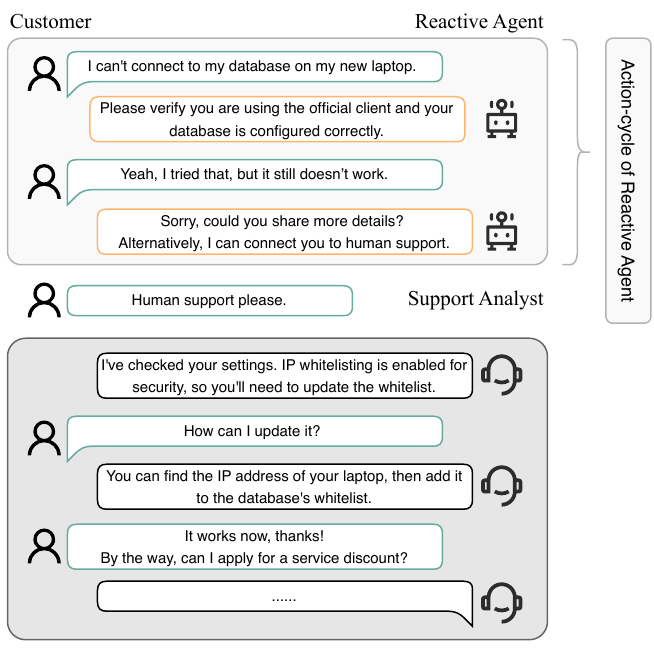}
        \label{fig:intro_reactive}
    }
    \hfill
    \subfigure[Proactive Agent: Identifies and selectively answers queries to assist human analysts.]{
        \includegraphics[width=.52\textwidth]{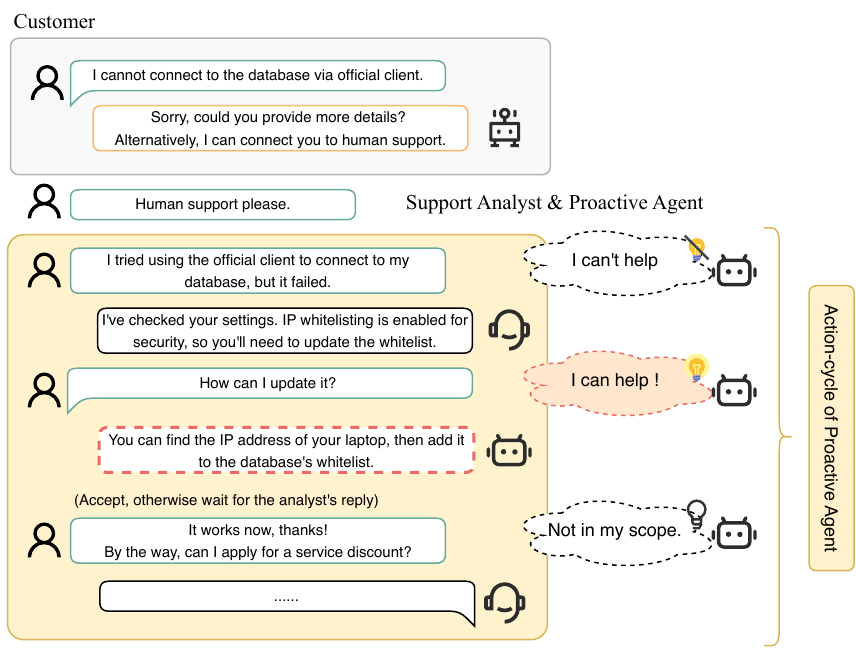}
        \label{fig:intro_proactive}
    }
    \caption{Comparison of reactive and proactive agent paradigms in the on-call support process. \normalfont{(a) Reactive agents disengage upon escalation, whereas (b) proactive agents extend the action-cycle to collaborate with human support analysts.}}
    \label{fig:intro}
\end{figure*}

Operational management of Volcano Engine, the large-scale public cloud platform of ByteDance, presents a significant challenge due to the high volume of on-call support requests\cite{liuTickItLeveragingLarge2025}.
The platform receives thousands of customer support tickets daily, which are primarily addressed through on-call chat dialogues.
This continuous influx of on-calls places considerable strain on human support analysts and engineers, who must handle urgent incidents while also sustaining feature development and routine operations.
When the on-call workload spikes (e.g., during large-scale service degradations or traffic bursts), timely human responses can become a bottleneck, and customers may experience prolonged waiting time throughout the entire on-call process.

To mitigate this operational burden, recent research has explored Large Language Models (LLMs)\cite{IntroducingGPT52025,googleGemini25Pro} to develop reactive agents for on-call assistance.
These reactive agents typically serve as the first line of support, interacting directly with customers to resolve initial inquiries.
However, their operational scope is often restricted, they are designed to disengage from the session once an inquiry exceeds their capabilities and is escalated to a human analyst.
Although prior work has demonstrated their effectiveness in specific scenarios\cite{arikkatIntellBotRetrievalAugmented2024,shafeeEvaluationLLMChatbots2025}, we identify two critical gaps that limit their utility for real-world on-call support in cloud platforms.

First, customers often initiate an on-call session with a high-level problem description, followed by a sequence of increasingly specific questions to converge on a resolution.
Progress in such sessions therefore depends on addressing these sub-questions sequentially. 
In the reactive paradigm, as illustrated in Figure \ref{fig:intro_reactive}, the agent is explicitly invoked by the customer and is expected to respond turn by turn, regardless of whether it can provide an adequate answer at each step.
However, once the agent fails to resolve a single sub-question, the session is typically escalated to a human analyst, and the customer’s progress depends on analyst availability.
After this handoff, the agent’s action cycle terminates, preventing it from assisting with subsequent sub-questions, including those well within its capabilities.
This premature disengagement leads to missed opportunities to further reduce analyst workload and shorten end-to-end response time.

A second significant challenge in on-call scenarios is the dynamic context and rapidly evolving knowledge.
While Retrieval-Augmented Generation (RAG)\cite{xuRetrievalAugmentedGenerationKnowledge2024,lewisRetrievalAugmentedGenerationKnowledgeIntensive2020} can improve an agent’s access to domain knowledge, they typically rely on a manually curated knowledge base\cite{kimKGGPTGeneralFramework2023}.
In practice, such knowledge bases are often relatively static and require continuous manual maintenance.
Consequently, the effectiveness of existing methods\cite{yangRAGVAEngineeringRetrieval2025,akkirajuFACTSBuildingRetrieval2024} depends heavily on the quality, coverage, and timeliness of curated content.
This mismatch is particularly costly in fast-evolving on-call situations, where newly discovered symptoms, mitigations, or temporary workarounds are often shared in active on-call sessions. 
Human analysts may also need to repeat the same explanations across multiple sessions or across different on-call shifts.
Because these insights are rarely incorporated into the knowledge base in a timely manner, the agent cannot leverage them to assist other customers experiencing the same issues, resulting in delayed responses and duplicated troubleshooting efforts. 
For instance, during a network service failure, critical information or temporary workarounds may first emerge in an existing on-call session, yet remain unavailable to the agent when similar tickets arrive shortly afterward.

To address these critical gaps and reduce the operational overhead of on-call support, we propose \name{}, a novel system that introduces a proactive agent to operate alongside human analysts throughout the entire on-call life-cycle.
Rather than replacing the first-line reactive agent, \name{} specifically targets the collaborative phase after escalation to human support, where analyst availability is often the bottleneck to the customer waiting time.
\name{} integrates directly into the dialogue between support analysts and customers, acting not as a sole interlocutor but as an assistant to the human analyst.
As illustrated in Figure \ref{fig:intro_proactive}, it continuously monitors the ongoing on-call dialogue, and whenever it identifies a customer question within its capabilities, it proactively provides an answer without requiring explicit invocation.
By remaining active after escalation, \name{} complements existing first-line reactive agents by extending its action-cycle beyond the initial triage stage, enabling proactive assistance throughout the on-call life-cycle and reducing analyst effort.

Furthermore, \name{} incorporates an automated, continuous self-improvement mechanism.
Our observations from real-world on-calls suggest that many questions, symptoms, and effective mitigations recur across different sessions, making knowledge captured in one incident valuable for subsequent ones.
By leveraging its extended action-cycle, \name{} can learn from both answered and unanswered questions, as well as additional external documents shared during the on-call conversations, turning these information into reusable knowledge.
Over time, this mechanism helps keep \name{}’s knowledge more up to date and improves answer quality, reducing repeated troubleshooting effort in future on-calls.

The salient contributions of our work are as follows:

\begin{itemize}
  \item We present \name{}, a proactive agent that \textbf{complements} first-line reactive agents by staying active after escalation and autonomously answering in-scope questions during analyst--customer dialogues, thereby extending the on-call action cycle.
  \item We design a \textbf{continuous self-improvement} mechanism that learns from on-call interactions and shared artifacts to update the agent's knowledge, reducing dependence on manually curated knowledge bases.
  \item We deploy \name{} in the Volcano Engine production environment for over ten months and evaluate it with quantitative metrics and case studies, demonstrating improved operational efficiency and on-call support quality.
\end{itemize}


\section{Related Work}
\label{sec:related}

The increasing volume of customer support tickets poses significant challenges for Volcano Engine\cite{liuTickItLeveragingLarge2025}.
To address this, our work explores the use of LLM-powered agents for on-call support, with the aim of improving efficiency.
This section reviews relevant studies on the application of LLMs in customer support, then introduces proactive agent paradigms, and finally studies the continuous self-improvement mechanism that enhance agent performance.

\textbf{LLM-based Agents in On-call Support.}
Recent advances have demonstrated that LLM-based agents possess remarkable capabilities in natural language understanding and complex reasoning.
The advent of LLMs has catalyzed a paradigm shift in automated customer support\cite{suLLMFriendlyKnowledgeRepresentation2025, shiCHOPSCHatCustOmer2024}.
A growing body of literature has focused on conversational agents\cite{huangCanChatbotCustomer2024,romeAskMeAnything2024}, or chatbots\cite{bhattacharyyaStudyAdoptionArtificial2024}, designed to handle customer inquiries and alleviate the workload of human support analysts.
Most existing work, however, adopts a reactive paradigm\cite{peiFlowofActionSOPEnhanced2025}, where the agent serves as the primary interlocutor with the customer \cite{huangCanChatbotCustomer2024,romeAskMeAnything2024}.
Such agents are often enhanced with external tools\cite{bhattacharyyaStudyAdoptionArtificial2024,patilGorillaLargeLanguage2024} and knowledge bases\cite{romeAskMeAnything2024, xuRetrievalAugmentedGenerationKnowledge2024, kimKGGPTGeneralFramework2023} to improve the coverage and accuracy.
Despite their effectiveness in specific scenarios, reactive agents remain limited by the requirement for explicit user invocation for each query\cite{wangEComBenchCanLLM2025}.
Their action cycle typically terminates if they fail to answer any question, at which point the customer must turn to human support analysts for assistance.
This reliance on a reactive action-cycle constrains their potential and underutilizes the broader capabilities of LLMs as compared to proactive approaches\cite{dengSurveyProactiveDialogue2023,dengHumancenteredProactiveConversational2024}.

\textbf{Proactive Agent Paradigms.}
The LLM-powered proactive agents are characterized by their ability to take initiative, acting without explicit user invocation.
Rather than merely responding to explicit requests, they infer user needs from contextual cues\cite{dengHumancenteredProactiveConversational2024}.
Recent studies have begun to explore proactive behaviors in various domains, including code completion\cite{chenEvaluatingLargeLanguage2021}, tabular data interpretation\cite{dengPACIFICProactiveConversational2022}, personalized social bots\cite{niuPaRTEnhancingProactive2025}, and general task assistance\cite{luProactiveAgentShifting2024, yangContextAgentContextAwareProactive2025}.
Further work has examined methods to strengthen proactivity through explicit planning\cite{zhangAskbeforePlanProactiveLanguage2024}, internal reasoning\cite{liuProactiveConversationalAgents2025}, and multi-agent collaboration\cite{zhangProAgentBuildingProactive2024}.
Nevertheless, the application of a proactive paradigm in on-call support remains underexplored.
Our work addresses this gap by introducing an agent that continuously analyzes an on-call dialogue and intervenes only when identifying an opportunity to assist.

\textbf{Continuous Self-Improvement Mechanism.}
A persistent challenge in deploying LLM agents in dynamic domains such as cloud services is ensuring that their knowledge remains up to date. 
To address this, recent research has investigated \emph{self-evolving} approaches \cite{gaoSurveySelfEvolvingAgents2025}.
Parametric methods, such as reinforcement learning-based adaptation \cite{zhouSelfChallengingLanguageModel2025}, can improve agent performance but require costly offline training, making them difficult to deploy in rapidly changing on-call contexts.
By contrast, non-parametric approaches \cite{chengLiftYourselfRetrievalaugmented2023, wangVoyagerOpenEndedEmbodied2023}, which avoid direct parameter fine-tuning, offer a more flexible alternative.
However, standard RAG\cite{lewisRetrievalAugmentedGenerationKnowledgeIntensive2020, xuRetrievalAugmentedGenerationKnowledge2024} typically relies on a knowledge base that is curated and updated manually.
This maintenance process introduces delays between the emergence of new knowledge(e.g., a service outage) and its availability to the agent.
Some methods\cite{chengLiftYourselfRetrievalaugmented2023, wangVoyagerOpenEndedEmbodied2023} enable agents to incrementally expand their knowledge bases from interactions and experiences, making them particularly well-suited for the dynamic requirements of on-call support.
In this work, we design a continuous self-improvement mechanism that enables the agent to learn from each proactive action, thereby autonomously enriching its knowledge base.

\begin{figure*}[t]
    \centering
    \includegraphics[width=\textwidth]{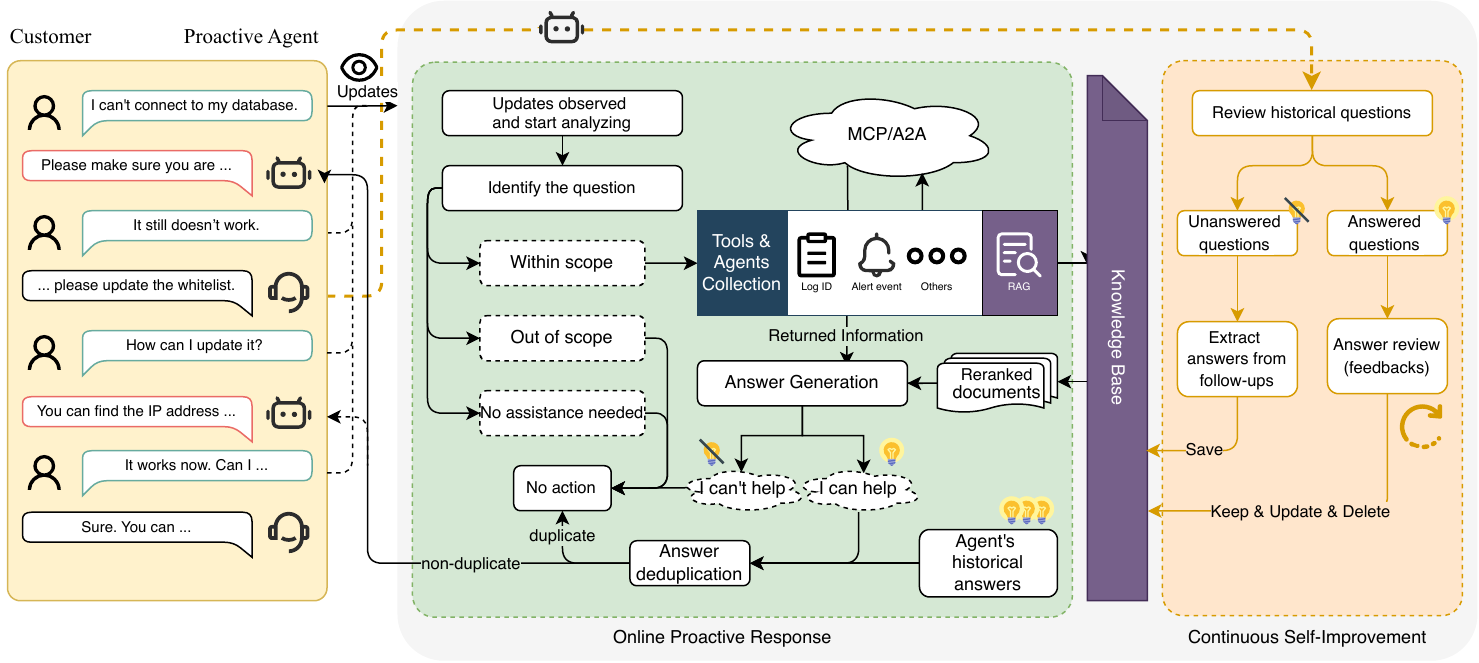}
    \caption{Framework of \name{}. \normalfont Features with two primary functions: (1) Online Proactive Response; (2) Continuous Self-Improvement.}
    \label{fig:framework}
\end{figure*}

\section{Methodology}
\label{sec:methodology}

In this section, we introduce \name{}, a proactive agent system designed for on-call support.
As illustrated in Figure \ref{fig:framework}, \name{} targets the collaborative phase after human intervention and operates as a proactive assistant within the dialogue between the customer and the human support analyst.
It performs two primary functions:
(1) \textbf{Online Proactive Response:} \name{} continuously monitors the on-call session, identifies questions, and proactively provides answers.
(2) \textbf{Continuous Self-Improvement:} \name{} reviews its proactive actions and learns from the on-call dialogue in a continuous manner, refining its knowledge base and improving future performance.

\subsection{Online Proactive Response}

Different from reactive agents that wait for explicit user invocation, \name{} proactively participates in the on-call between the customer and the human analyst, responding at opportune moments.
The action-cycle for \name{} initiates when a human support analyst intervenes in the on-call session and concludes upon session closure.
The proactive response mechanism involves three key tasks.

\subsubsection{Context-Aware Question Identification}

To function effectively as a complementary assistant alongside human experts, \name{} must discern \textit{when} to intervene before \textit{what} to answer. 
The identification module is designed to bridge the response gap by targeting moments where a customer raises a valid sub-question that the human analyst has not yet addressed, often occurring while the human analyst is occupied with complex diagnostics or multitasking.

A critical prerequisite for this identification is a clear capability scope of the agent.
\name{} is specialized for the cloud service platform, with its primary mission being the resolution of inquiries regarding cloud products, services, and diverse operational issues.
This clear and focused scope ensures that the agent remains dedicated to the platform's domain knowledge rather than behaving as a generic question-answering assistant.
When the on-call dialogue is updated with new messages from the customer, \name{} identifies whether the current message falls within this defined scope.

Formally, this identification task is framed as a classification problem. 
As Figure \ref{fig:framework} shows, \name{} analyzes the latest message to categorize it into the following three classes:

\textbf{Within Scope.} The message contains a specific question within the scope of \name{} agent, regarding the cloud platform's products or operational issues.

\textbf{Out of Scope.} The message contains questions beyond the specialized capabilities of the agent, such as complex subjective decision-making best left to human judgment.

\textbf{No assistance needed.} The message consists of phatic communication, such as greetings, affirmations or queries that the human analyst has already explicitly acknowledged or resolved.

Accurate scope confirmation is paramount to maintain user trust and avoid intrusive or irrelevant interruptions.
We implement this identification logic using a specialized prompt for the LLM, the details of which are provided in Appendix \ref{appendix:question_identification}.
\name{} proceeds to the answer generation phase only when a message is classified as \textit{Within Scope}, ensuring its proactive response is both timely and necessary.
For questions outside this scope, \name{} is expected to refrain from intervening.

Furthermore, for some operational issues, \name{}'s capability extends by invoking tools to retrieve detailed logs, associated alerts, and diagnostic metadata. 
This integration allows \name{} to ground its responses in real cloud system, providing highly specific and actionable assistance.

\subsubsection{Answer Generation}

After identifying a question within its capability scope, \name{} proceeds to generate an answer.
Because the on-call sessions are multi-turn dialogues, customer questions may be implicit and require a comprehensive understanding of the conversational context.
Therefore, \name{} first analyzes the entire dialogue history to rewrite the newly identified question.
This step utilizes established techniques\cite{maoRaFeRankingFeedback2024,maQueryRewritingRetrievalAugmented2023} to make the question explicit and self-contained, including deconstructing complex queries and resolving coreferences and ellipses.

Next, the rewritten questions are processed by the agent.
Retrieval-Augmented Generation (RAG) plays a crucial role in this process, not only supplementing domain-specific knowledge gaps in the foundational LLM but also providing reliable references for the generated answers.
Specifically, \name{} employs a multi-path knowledge retrieval strategy\cite{bestaMultiHeadRAGSolving2025,wangRichRAGCraftingRich2024}, which includes a domain-related document knowledge base and a collection of question-answer pairs from historical on-call sessions.
A reranking model then reorders the retrieved results based on their relevance to the question.

Finally, \name{} synthesizes all collected information, including the multi-turn dialogue context, tools-calling information and retrieved knowledge, to generate a final answer.
The agent retains the option to refuse to answer if it determines that the available information is insufficient for a reliable response.
A detailed prompt for this is in Appendix \ref{appendix: answer_generation}.

\subsubsection{Answer Deduplication}

In a real-world on-call session, the customer may repeatedly discuss a same topic by rephrasing the question, providing additional background, or expressing dissatisfaction with a previous answer.
The proactive mechanism of \name{} could trigger multiple and similar responses to these updates.
To avoid this, \name{} performs an answer deduplication step to intercept redundant replies.

We employ a semantic similarity-based deduplication method for this task.
When a new answer is generated, its embedding vector is calculated using a pre-trained sentence embedding model.
The cosine similarity of this embedded vector is then computed against the embeddings of all historical answers sent during the current on-call session.
If the maximum similarity exceeds a predefined threshold $\theta$, the new answer is considered redundant and suppressed.
The threshold $\theta$ is set empirically to balance the trade-off between redundancy avoidance with comprehensive assistance.
It is highly relevant to the embedding model, and we use an empirical value of 0.7, which is further discussed in the evaluation section.

If the answer is considered to be unique, \name{} sends an answer card to the on-call.
Compared to LLM-based approaches, this similarity-based deduplication method is significantly faster while remaining effective.
We perform deduplication at the answer stage rather than the question identification stage, because the knowledge base may be updated as the on-call session progresses.
Consequently, even identical may lead to different responses within the same session, making answer-level deduplication more appropriate.

\subsubsection{Proactive Response Card Design}
\label{subsubsec:card_design}

\begin{figure}[t]
    \centering
    \includegraphics[width=\linewidth]{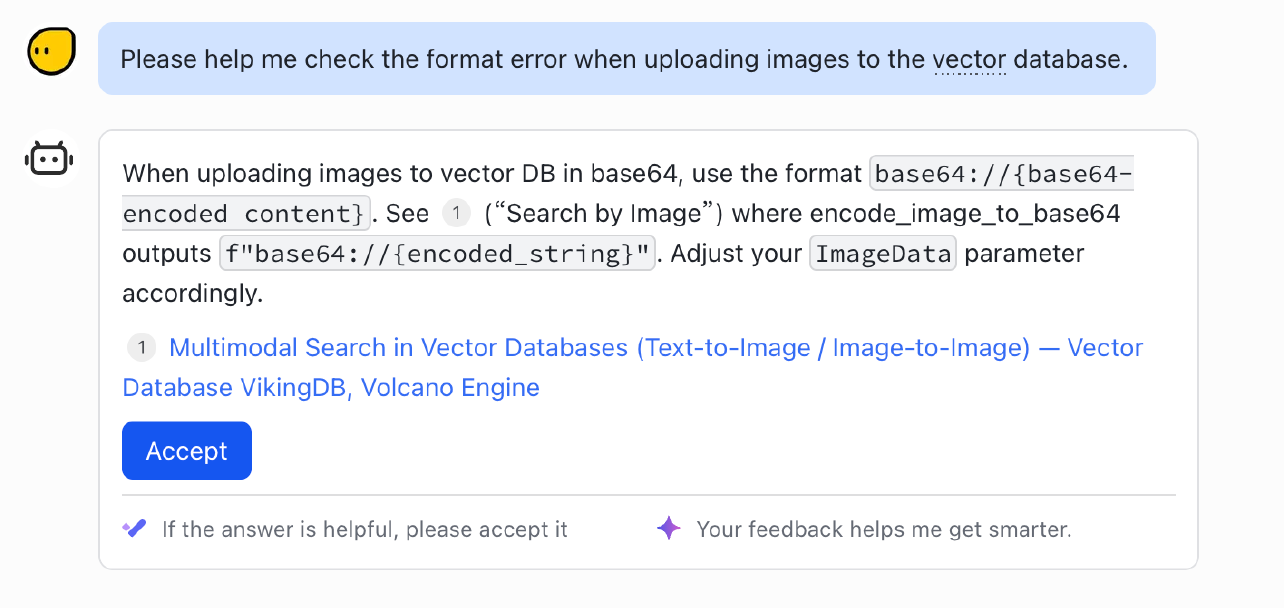}
    \caption{Proactive response card of \name{}. \normalfont Including a distinct layout for distinguishing agent from human, explicit citations for verifiability and an \textit{Accept} button for collecting feedback.}
    \label{fig:answered_question_demo}
\end{figure}

To ensure that the proactive responses of \name{} are constructive without being intrusive or misleading, we employ a specialized card-based interface. 
Unlike reactive agents that often dominate the conversational flow, \name{} presents its findings in a structured card that visually demarcates LLM-generated content from the ongoing human-to-human dialogue.

As shown in Figure \ref{fig:answered_question_demo}, this card features a distinct avatar and layout to differentiate it from human messages.
To ensure ethical transparency, the card explicitly displays the unique identity and provides citation links to ground the generated answer.
This clear distinction ensures that customers can instantly discern automated suggestions from human analyst responses.

Furthermore, the card incorporates an interactive \textit{Accept} button for feedback.
It allows customers to acknowledge the answer or human analysts to verify it, effectively marking the question as resolved to streamline the on-call process.
Simultaneously, this interaction serves as verified ground truth, helping \name{} assimilate high-quality answers for self-improvement.

\subsection{Continuous Self-Improvement}

Benefiting from the proactive mechanism, \name{} maintains a long action-cycle that spans the entire on-call session.
It not only answers the questions, but also follows the dialogue updates from both the customer and the human analyst.
This capability allows \name{} to learn from customer feedback and support analyst responses, enabling continuous refinement of its knowledge base.
As shown in Figure \ref{fig:answer_review}, this module reviews every query that within its scope, handling answered and unanswered questions differently.

\subsubsection{Learning from Unanswered Questions}

Even for questions within its capability scope, \name{} may occasionally fail to generate a response due to the absence of suitable references in the existing knowledge base. 
In these scenarios, the human support analyst inevitably steps in to resolve the issue.
The mandatory nature of on-call support guarantees that human analysts must eventually address these customer inquiries.
\name{} treats these human-led resolutions as valuable learning opportunities.

\name{} is designed to autonomously parse the subsequent analyst-customer dialogue and extract the solution provided by the human expert.
If a definitive answer is identified, it is structured as a new question-answer pair and stored into the knowledge base. 
This process incrementally enriches the knowledge base and expands the coverage for future inquiries, enabling \name{} to handle similar queries in the future.

It is important to note that these autonomously extracted answers may initially be constrained by the specific context of the original dialogue. 
However, this limitation is temporary. 
Once these entries are retrieved and utilized in future on-calls, they become subject to the validation mechanisms in the \textit{Learning from Answered Questions} module. 
Through the Update operation described previously, these specific answers are iteratively polished and generalized, ensuring the continuous evolution of the knowledge base quality.

\subsubsection{Learning from Answered Questions}
\label{subsubsection:learning_from_answered_questions}

For the questions that \name{} can address, it proactively sends an answer card to the on-call.
We consider positive feedback (e.g., an \textit{Accept} click) as a strong endorsement. 
Accordingly, the corresponding rewritten question and generated answer can be stored as a new question-answer (QA) pair in the knowledge base.

Notably, a direct negative feedback option is omitted.
This design is motivated by the observation that the criteria for negative feedback in real-world dialogues are often ambiguous and inconsistent.
For instance, users may flag an answer as unsatisfactory due to minor factual inaccuracies, incomplete explanations, irrelevant references, or broken reference links.
In other cases, users might simply skip providing feedback and continue the on-call dialogue, even when they are dissatisfied.
Such variability makes it difficult to reliably interpret negative feedback and collect the detailed reasons behind it.

To ensure a more consistent refinement process, \name{} employs an agent to automatically review and evaluate answers that appear to be unaccepted.
\name{} identifies whether a human analyst subsequently provides a more satisfactory answer (Appendix \ref{appendix:answer_review}) and takes action accordingly.
As detailed in Figure \ref{fig:answer_review}, the review process follows the following three paths:

\begin{figure}[t]
    \centering
    \includegraphics[width=\linewidth]{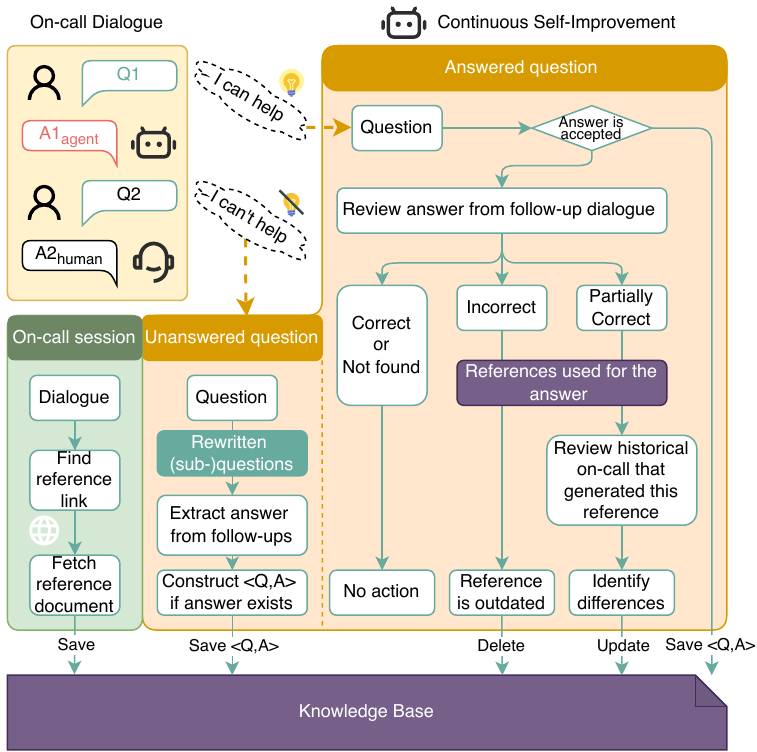}
    \caption{Continuous self-improvement framework.
    \normalfont \name{} learns from answered questions, unanswered questions, and external documents shared in the on-call dialogue.}
    \label{fig:answer_review}
\end{figure}

\textbf{Keep.}
If the human support analyst provides a follow-up answer consistent with the answer generated by \name{}, or if the dialogue proceeds without further discussion of the topic, the answer is regarded as correct.
In these cases, the reference used to generate the answer is deemed appropriate for the question, and no updates is required.
It is important to note that this \textit{Keep} status is not permanent, every retained entry remains subject to continuous validation in future proactive cycles, ensuring that knowledge which becomes obsolete is eventually re-evaluated.

\textbf{Delete.}
In some cases, the review agent finds that an answer is completely incorrect based on the follow-up on-call dialogue.
We note that the heavily depended references, which were extracted from previous on-call sessions, may not match the context of the current question.
For instance, the references may be outdated, too specific to generalize, or no longer suitable for the question.
To prevent these inappropriate references from misleading future answers, \name{} deletes them from the knowledge base.

\textbf{Update.}
If the answer has only minor discrepancies compared to the subsequent human-provided answer, the agent reviews the causes of these differences.
Such discrepancies may arise from variations in background context or missing prerequisite information.
To address this, \name{} augments the references with additional details that help identify the scenario of the question and improve answer accuracy.
The refined question–answer pair is then used to update the corresponding entry in the knowledge base.

Overall, this review-action framework establishes a robust self-correcting feedback loop. 
Even if a temporary workaround or a non-generic solution is autonomously learned from a human analyst, its long-term impact is mitigated by the system's proactive nature. 
If such an entry leads to an unaccepted answer in future sessions, it inevitably triggers the \textit{Update} or \textit{Delete} process during the next review cycle. 
This iterative verification effectively prevents performance degradation caused by outdated or low-quality knowledge.

\subsubsection{Learning from External Documents}

In addition to learning from direct dialogues, \name{} actively reviews external documents referenced within on-call sessions to extend its learning scope. 
This mechanism mirrors the workflow of human analysts, who frequently validate their solutions by sharing links to official documentation, standard operating procedures, or API references.

This reliance on external resources addresses two critical challenges in textual support. 
First, for intricate issues requiring sequential execution, typing out lengthy instructions in a chat interface is often inefficient and prone to ambiguity; comprehensive manuals provide a superior, structured format. 
Second, for dynamic content such as API parameters and versioning, static text in chat messages risks becoming rapidly obsolete. 
By linking to official documents, analysts ensure customers access the most current and canonical information.

Unlike the fragmented information found in casual conversation, these documents offer systematic and reusable knowledge. 
\name{} is designed to automatically capture these high-value resources by parsing the shared links. 
By extracting relevant content from these authoritative sources, the agent transforms transient support interactions into a persistent, multi-faceted knowledge repository, ensuring its answers are accurate, up-to-date, and well-documented.

\section{Experiments}
\label{sec:experiments}

This section presents a comprehensive evaluation of \name{}, designed to answer the following research questions:

Q1: To what extent does \name{} provide extended support coverage during the human-involved phase of on-call sessions?

Q2: How accurately can \name{} identify questions that fall within its ability scope?

Q3: How accurately can \name{} proactively provide answers to the questions it identifies?

Q4: Can \name{} effectively prevent providing duplicate answers within the same on-call dialogue?

\subsection{Real-world Deployment and Statistics}

The \name{} system has been deployed in the production environment of the Volcano Engine cloud platform since March 13, 2025.
The deployment was phased, beginning with a limited set of on-calls and gradually expanding to full coverage by June 1, 2025.
By January 16, 2026, \name{} had processed a total of 131,433 on-calls, encompassing 2,317,760 dialogue messages. 
During this period, \name{} autonomously collected 272,325 knowledge entries and generated 136,234 proactive responses.

\subsection{Dataset Construction and Labeling}

\subsubsection{Data Sampling}
\label{subsubsec:data_sampling}
To conduct a rigorous evaluation of \name{} for the research questions Q2, Q3, and Q4, we constructed a high-quality evaluation dataset.
It is important to note that real-world on-call inquiries are inherently multi-modal, typically consisting of textual descriptions accompanied by visual evidence, such as screenshots of error logs or monitoring dashboards.
To reflect this reality, our sampling process preserved both text and image data.

Considering that the production environment involves sensitive customer data, and some foundation models used in our experiments are closed-source (e.g., GPT and Gemini series models), we sample 300 non-sensitive on-call data from the production environment.
All subsequent experiments are strictly conducted on the 300 non-sensitive on-call datasets to ensure data security compliance.

\subsubsection{Data Labeling}

To establish a robust ground truth, we employed a hybrid labeling strategy that combines explicit online feedback with offline expert review.

As described in Section \ref{subsubsec:card_design}, the proactive response card features an interactive \textit{Accept} button (Figure \ref{fig:answered_question_demo}).
We treat this interaction as a high-confidence signal for automatic labeling.
Specifically, if a response is explicitly accepted during the on-call session, it is automatically labeled as a positive sample for scope identification and a correct answer for generation quality.
This online mechanism ensures that a portion of our ground truth directly reflects real-world utility and user satisfaction.

However, relying solely on online feedback may yield false negatives, as accurate responses might be inadvertently overlooked.
To address this, we engaged human support analysts to refine the dataset for the 300 sampled on-calls.
The analysts performed a comprehensive review to handle the remaining data.
For each message from the on-call customers, human analysts first manually label whether the message is within the scope of \name{}, such as whether the message contains a question related to the platform.
This can help to evaluate the question identify module of \name{}, making it seize every opportunity to assist.
Besides, for generated answers that without being accepted, analysts evaluate their accuracy, labeling them as \textit{Correct} or \textit{Incorrect}.
This can help further verify the correctness of these answers, thereby determining whether these responses are helpful and providing a positive impact on the overall on-call process.

\subsection{Experiment Settings}

\subsubsection{Foundation Models Comparison}
As a generic system, \name{} is designed to be compatible with various foundation large language models.
Given the multi-modal nature of the dataset constructed in Section \ref{subsubsec:data_sampling}, it is imperative to utilize models capable of processing both text and visual inputs.
Therefore, to evaluate the flexibility and performance of \name{}, we select a diverse set of state-of-the-art multi-modal LLMs as the foundation models in this experiments.
The chosen models range from top-tier proprietary models, including Seed-1.6, Seed-1.6-flash\cite{seedByteDanceSeed}, GPT-5, GPT-5-mini, Gemini-2.5-pro, Gemini-2.5-flash\cite{googleGemini25Pro}, to the open-weight model Qwen-2.5-VL-72b\cite{qwenQwen25VL}.

It is worth noting that in real-world industrial deployments, the selection of a foundation model is rarely dictated by generation quality alone.
Critical factors such as data security, compliance, inference latency, and operational cost often take precedence.
For instance, while proprietary SOTA models may offer superior reasoning, internal security policies may mandate the use of self-hosted or region-specific models for handling sensitive customer data.
Although we do not explicitly model these constraints in our experiments, we account for this industrial reality by ensuring \name{} is tested across a broad spectrum of models.
Consequently, our focus is not on identifying the best foundation model, but on assessing the effectiveness of \name{} with different models via ablation and parameter studies across this diverse landscape.

\subsubsection{Evaluation Metrics}
The experimental evaluation is conducted using $Accuracy$ ($\frac{TP+TN}{TP+TN+FP+FN}$), $Precision$ ($\frac{TP}{TP+FP}$) and $Recall$ ($\frac{TP}{TP+FN}$).
To account for the class imbalance present in the dataset, we further report the weighted precision ($Precision_{w}=\frac{\sum n_iPrecision_i}{\sum n_i}$)), weighted recall ($Recall_{w}=\frac{\sum n_iRecall_i}{\sum n_i}$), and weighted F1-score ($F1_{w} = \frac{\sum n_i F1}{\sum n_i}$) as additional evaluation metrics.

Given the diverse constraints that dictate model selection in production, comparing the absolute scores of different foundation models provides limited insight into the \name{} framework itself.
Therefore, in addition to absolute scores, we highlight the relative improvement (marked as $\uparrow$) that \name{} achieves for each specific foundation model, as this marginal utility is the key indicator of the effectiveness of \name{} in production.

To align with the evaluation targets of \name{}, we adopt task-specific evaluation protocols. 
For scope identification (Q2) of \name{}, we directly compute the aforementioned classification metrics against the human-annotated ground truth. 
For answer correctness (Q3), we employ an LLM-as-a-Judge\cite{guSurveyLLMasaJudge2025} to determine whether a generated response is factually correct based on the user message and associated visual evidence. 
The judge’s binary decision is then used to calculate the same set of metrics.

\begin{figure}[t]
    \centering
    \subfigure[Distribution of On-calls by Volume of Extended Assistance Provided by \name{}]{
        \includegraphics[width=\linewidth]{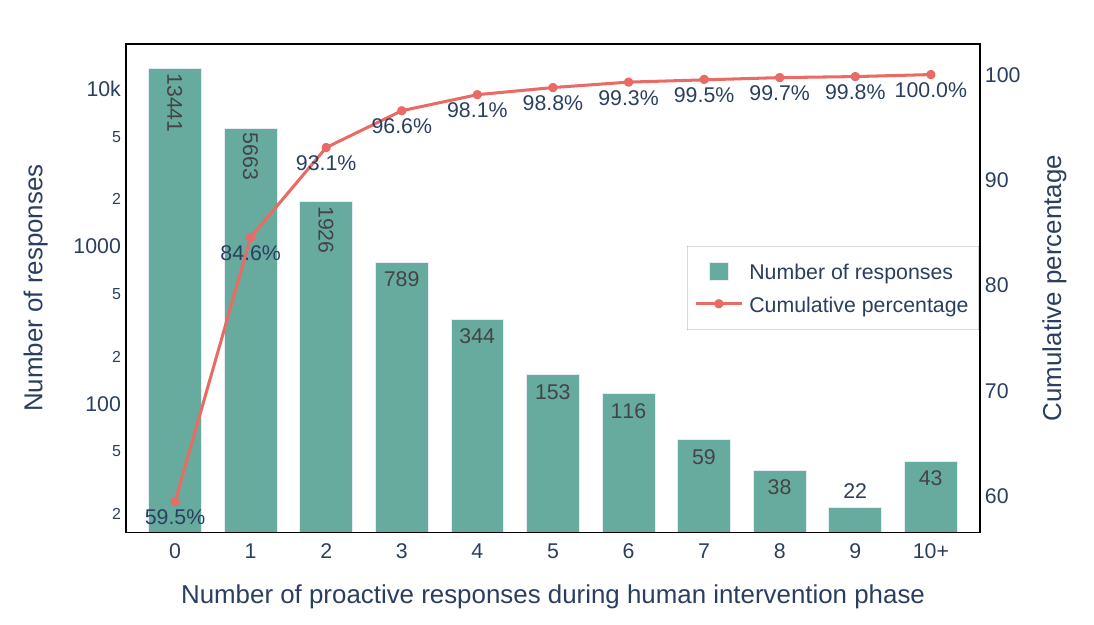}
        \label{fig:response_cnt}
    }
    \subfigure[Number of Proactive Responses of \name{} Over Time]{
        \includegraphics[width=\linewidth]{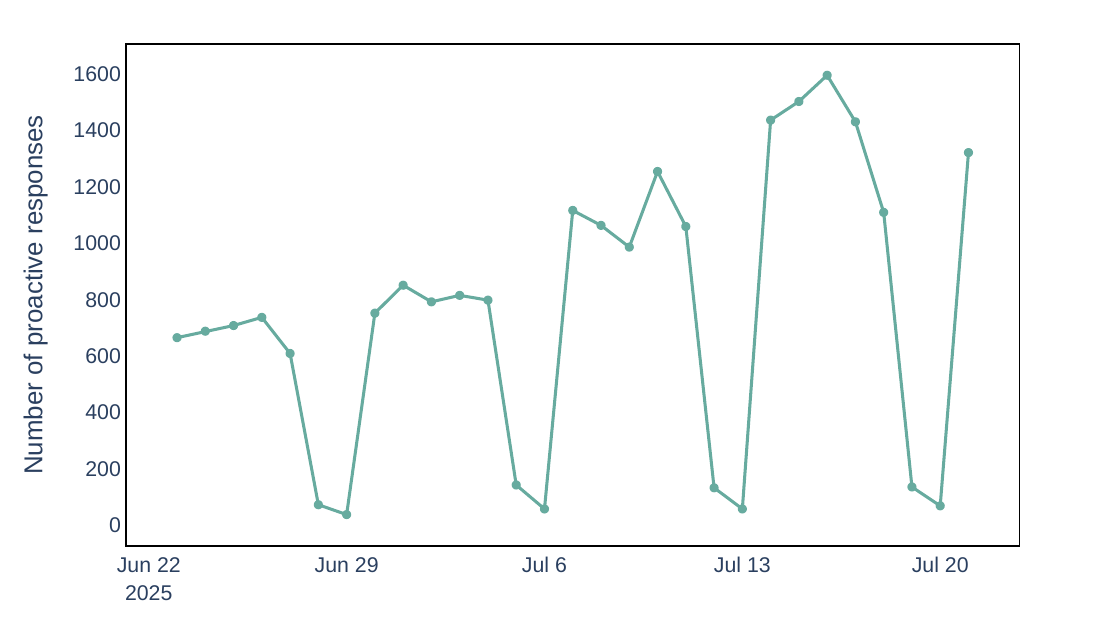}
        \label{fig:response_by_date}
    }
    \caption{Volcano Engine On-Call Statistics, June 23 – July 22, 2025}
    \label{fig:deduplication}
\end{figure}

\begin{table*}[t]

    \caption{Evaluation of question identification of \name{} across various foundation models.}
    \centering
    \begin{tabular}{c|ccc|ccc|ccc}
    \toprule
    Foundation
& \multicolumn{3}{c|}{$Precision_w$}
& \multicolumn{3}{c|}{$Recall_w$/Accuracy}
& \multicolumn{3}{c}{$F1_w$} \\
Model
& Baseline  & \name{} & $\Delta$ (\%)
& Baseline  & \name{} & $\Delta$ (\%)
& Baseline  & \name{} & $\Delta$ (\%)  \\
    \midrule
        GPT-5 & 0.789 & 0.870 & 10.3\% & 0.620 & 0.850 & 37.1\% & 0.660 & 0.857 & 29.8\% \\
        GPT-5-mini & 0.849 & 0.875 & 3.0\% & 0.307 & 0.793 & 158.7\% & 0.268 & 0.813 & 203.2\% \\
        Gemini-2.5-flash & 0.811 & 0.859 & 6.0\% & 0.700 & 0.737 & 5.2\% & 0.730 & 0.764 & 4.6\% \\
        Gemini-2.5-pro & 0.784 & 0.827 & 5.5\% & 0.757 & 0.830 & 9.7\% & 0.768 & 0.828 & 7.9\% \\
        QWen-VL-72b & 0.848 & 0.827 & -2.5\% & 0.287 & 0.677 & 136.0\% & 0.235 & 0.711 & 202.0\% \\
        Seed-1.6 & 0.799 & 0.837 & 4.9\% & 0.640 & 0.813 & 27.1\% & 0.678 & 0.822 & 21.3\% \\
        Seed-1.6-flash & 0.805 & 0.842 & 4.6\% & 0.560 & 0.657 & 17.3\% & 0.601 & 0.692 & 15.2\% \\
    \bottomrule
    \end{tabular}

    \label{tab:question_identification}
\end{table*}

\subsection{Q1: Statistical Analysis of the Extended Action-Cycle}

We investigate Q1 by quantifying the additional support provided by \name{}.
Instead of replacing the first-line support which is often handled by reactive agents, \name{} is designed to extend the automated action-cycle into the human-intervention phase.
To analyze this, we conduct a statistical analysis on 22,594 on-call sessions over a one-month period from June 23 to July 22, 2025.

Figure \ref{fig:response_cnt} presents the distribution of on-calls based on the volume of responses provided by \name{} after human intervention.
As shows from the figure, in 40.5\% of all on-calls, \name{} continued to provide valuable assistance even after the human analyst had joined the session.
This result highlights the benefit of the extended action-cycle, rather than disengaging upon escalation, \name{} acts as a complementary assistant, addressing sub-questions and retrieving information alongside the human expert.

Furthermore, a key feature of this deployment is its zero-setup design, \name{} integrates into all on-calls without requiring a manually curated knowledge base.
It autonomously enriches its knowledge base from all the dialogues.
Figure \ref{fig:response_by_date} illustrates a consistent upward trend in the number of daily proactive responses, indicating that the agent is successfully accumulating knowledge from human-led resolutions to cover a broader range of queries over time.

\subsection{Q2: Question Identification Evaluation}

To evaluate the question identification module of \name{}, we compare the performance of several foundation LLMs with and without this module.
In the baseline configuration, which is without this module, the foundation model needs to implicitly decide whether to answer a query during the answer generation phase.
For this experiment, we regard on-call messages that should be answered as positive samples, and those that should not be answered as negative samples.
Performance is then measured using $Precision_w$, $Recall_w$ which is equal to $Accuracy$, and $F1_w$.

Table~\ref{tab:question_identification} indicates that the dedicated question identification module of \name{} improves performance across nearly all metrics and models.
This enhancement can be attributed to decomposing the complex task of simultaneously identifying and answering a question into a discrete classification step.
The degree of improvement varies among the models.
The Gemini series, which exhibits strong inherent capabilities in question identification, shows more modest gains.
The $F1_w$ increased by 4.6\% for Gemini-2.5-flash and 7.9\% for Gemini-2.5-pro.
Other models, including the GPT-5, Seed-1.6, and Qwen-VL series, demonstrate significant advancements when augmented with \name{}.
Overall, the improvement in the $F1_w$ ranges from 15.2\% for Seed-1.6-flash to 203.2\% for GPT-5-mini.
These results suggest that models initially less suitable for this task can achieve proficient performance with the introduction of the question identification module, such as QWen-VL-72b (202.0\% improvement) and GPT-5-mini (203.2\% improvement).
Furthermore, this module offers a flexible mechanism for constraining \name{} to address domain-specific inquiries, such as questions regarding cloud platforms, while filtering out general queries.

\subsection{Q3: Ablation Study on the Accuracy of Proactively Generated Answers}

\begin{table}[t]

    \caption{Ablation Study on the Answer Accuracy of \name{}}
    \centering
    \begin{tabular}{c|ccc}
    \toprule

Foundation

& \multicolumn{1}{c}{$w/o$}
& \multicolumn{1}{c}{$w/o$}
& \multicolumn{1}{c}{$w/$} \\
Model
& self-impr. & answer review & \name{} \\
    \midrule
GPT-5 & 0.389 & 0.537  (37.9\%$\uparrow$) & 0.624  (60.3\%$\uparrow$) \\
GPT-5-mini & 0.178 & 0.322  (81.1\%$\uparrow$) & 0.493  (177.4\%$\uparrow$) \\
Gemini-2.5-flash & 0.161 & 0.399  (147.9\%$\uparrow$) & 0.423  (162.5\%$\uparrow$) \\
Gemini-2.5-pro & 0.252 & 0.430  (70.7\%$\uparrow$) & 0.456  (81.3\%$\uparrow$) \\
Qwen-VL-72b & 0.225 & 0.322  (43.3\%$\uparrow$) & 0.396  (76.1\%$\uparrow$) \\
Seed-1.6 & 0.154 & 0.409  (165.2\%$\uparrow$) & 0.591  (282.6\%$\uparrow$)\\
Seed-1.6-flash & 0.158 & 0.379  (140.4\%$\uparrow$) & 0.510  (223.4\%$\uparrow$) \\
    \bottomrule
    \end{tabular}

    \label{tab:answer_accuracy}
\end{table}

The accuracy of proactively answered questions is a critical metric for evaluating the effectiveness of \name{}.
It reflects whether \name{} can provide correct and helpful information.
Following our evaluation protocol, we use an LLM-as-a-Judge\cite{guSurveyLLMasaJudge2025} to determine whether each proactively generated response is \textit{Correct} or \textit{Incorrect} given the user inquiry (incorporating associated visual evidence where available), and report accuracy based on these binary judgments.
We compared the full \name{} system against two ablated versions: one without the entire self-improvement module ($w/o$ self-impr.) and another without only the answer review component ($w/o$ answer review). The variant labeled "$w/o$ self-impr." relies solely on a foundation LLM and is treated as the baseline. 

The results in Table \ref{tab:answer_accuracy} show that the full \name{} system consistently and substantially outperforms both ablated versions across all models.
The self-improvement module provides the large gains by enabling the agent to learn domain-specific knowledge from dialogues autonomously.
With a minimum improvement of 60.3\% for GPT-5 and a maximum of 282.6\% for Seed-1.6.
The answer review component further refines accuracy by updating incomplete knowledge and deleting outdated entries.
This layered approach demonstrates that continuous learning is essential for improving answer accuracy over time.
In our real-world deployment which is based on Seed-1.6, the accuracy of proactively answered questions is 0.591.
In practice, this means that more than half of the questions identified and answered by \name{} are correct and helpful, significantly reducing the workload of human support analysts.

\subsection{Q4: Answer Deduplication Evaluation}

In real-world on-call scenarios, multi-turn conversations may revolve around the same topic.
In \name{}, a predefined threshold $\theta$ is introduced to balance the trade-off between avoiding redundancy and ensuring comprehensive assistance.
Noting that the optimal parameter may vary across different foundation embedding models, we conducted parameter selection experiments on a fixed embedding model seed-1.6-embedding\cite{seedSeed16Embedding}.

In this evaluation, we designate questions that are expected to be answered as positive samples, while those that should not be answered are treated as negative samples.
Under this condition, $Precision$ reflects the ability of the deduplication module to suppress redundant responses,
while $Recall$ measures its ability to preserve useful and distinct answers.
Considering that the imbalance of these two classes, we also report the weighted evaluation metrics.
By comparing these metrics under different threshold settings, we assess the effectiveness of our approach as Table \ref{tab:deduplication_threshold} shows.

\begin{table}[t]
    \centering
    \caption{Evaluation metrics of answer deduplication under different threshold $\theta$.}
    \begin{threeparttable}
    \begin{tabular}{c|c|c|c|c|c}
    \toprule
        $\theta$ & $Precision$ & $Recall$ & $Precision_{w}$ & $Recall_{w}$ & $F1_{w}$ \\
        \midrule
        0.00\tnote{a} & 1.000 & 0.396 & 0.864 & 0.509 & 0.542 \\
        0.20 & 0.996 & 0.398 & 0.861 & 0.510 & 0.543 \\
        0.40 & 0.886 & 0.585 & 0.771 & 0.602 & 0.646 \\
        0.60 & 0.826 & 0.921 & 0.730 & 0.778 & 0.747 \\
        \textbf{0.70} & 0.822 & 0.982 & 0.760 & 0.812 & \textbf{0.752} \\
        0.80 & 0.817 & 0.994 & 0.770 & 0.814 & 0.740 \\
        0.90 & 0.816 & 0.999 & 0.832 & 0.816 & 0.738 \\
        1.00\tnote{b} & 0.813 & 1.000 & 0.848 & 0.813 & 0.731 \\
    \bottomrule
    \end{tabular}
    \begin{tablenotes}
        \footnotesize
        \item[a] $\theta=0$ means all answers after the first are treated as duplicates.
        \item[b] $\theta=1$ corresponds to an ablation study of the answer deduplication module.
    \end{tablenotes}
    \end{threeparttable}
    \label{tab:deduplication_threshold}
\end{table}

In particular, setting the threshold to 0 ($\theta = 0$) means that after the first answer is sent, all the following answers will be considered as duplicates.
While setting the threshold to 1 ($\theta = 1$) corresponds to an ablation study of the answer deduplication module, where all answers are returned without deduplication.
From the results, we observe that as the threshold increases, the recall improves while precision declines.
The primary evaluation metric, the weighted F1-score ($F1_{w}$), is maximized at a threshold of 0.7, achieving a score of 0.752.
This result is highlighted as the optimal trade-off based on the weighted metrics.

\section{Case Studies}
\label{sec:case_studies}
To complement the quantitative evaluation, this section presents case studies from real-world deployments on the Volcano Engine cloud platform.
Two representative scenarios are examined to illustrate the practical advantages of \name{}.

Case 1: It demonstrates how the timely and automated self-improvement mechanism of \name{} enhances proactive response capabilities during emergency incidents.

Case 2: It illustrates the iterative evolution of knowledge, where \name{} identifies mismatched information and progressively refines the contextual precision of its knowledge base.

\subsection{Case1: Proactive Response Enhancement in Emergency Incidents}

\begin{figure}[t]
    \centering
    \includegraphics[width=\linewidth]{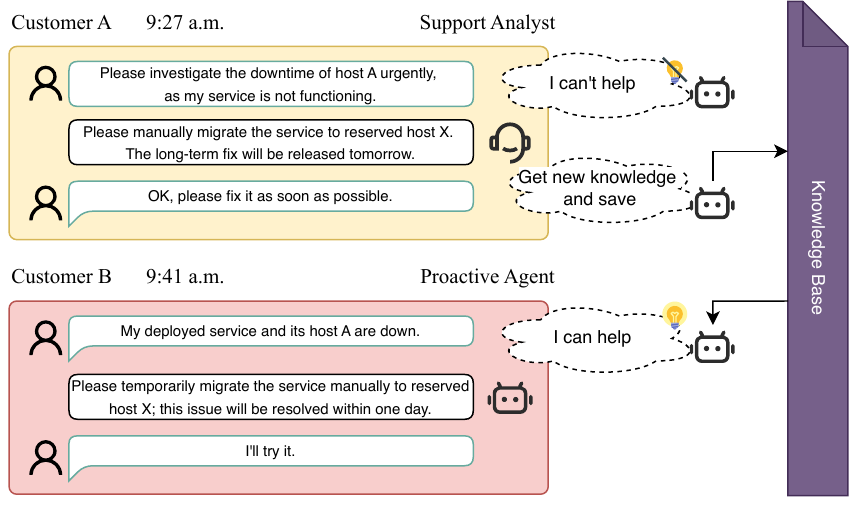}
    \caption{\name{} operates different on-call dialogues for the same critical issue (simplified and anonymized).}
    \label{fig:case1}
\end{figure}

The effective management of emergency incidents, though infrequent, is afforded the highest priority in Volcano Engine.
An emergency incident often manifests as multiple on-calls within a short period.
Critical information in such situations, including problem context, service status, troubleshooting progresses and proposed resolutions, are often fragmented across different on-call dialogues.

A notable incident occurred in June 2025, when different customer reported service disruption caused by the host machine failure within the same day.
As Figure \ref{fig:case1} shows, during the first on-call support, \name{} was unable to provide a resolution due to insufficient contextual information.
The site reliability engineering team observed that the automatic evacuation protocol had failed to execute as expected.
The team proposed a temporary workaround of manually migrating the service to a reserved host, while simultaneously preparing a long-term fix within a day.
By processing both the human analyst’s response and an associated support document from this initial incident, \name{} assimilated the necessary knowledge.
Later that day, another customer raised another on-call for the identical incident.
This time, benefiting from its automated self-improvement mechanism, \name{} proactively provided the correct temporary workaround before a human analyst intervened.

This example demonstrates how \name{} effectively reused knowledge from the first incident to proactively address the second one, thereby improving the efficiency.
This success is directly attributable to its continuous self-improvement framework, which enables timely learning from resolved inquiries and shared documentation.
Such automation significantly enhances technical support efficiency and delivers substantial value in emergency scenarios.

\subsection{Case2: Iterative Improvement of Response Quality}
\begin{figure}[t]
    \centering
    \includegraphics[width=\linewidth]{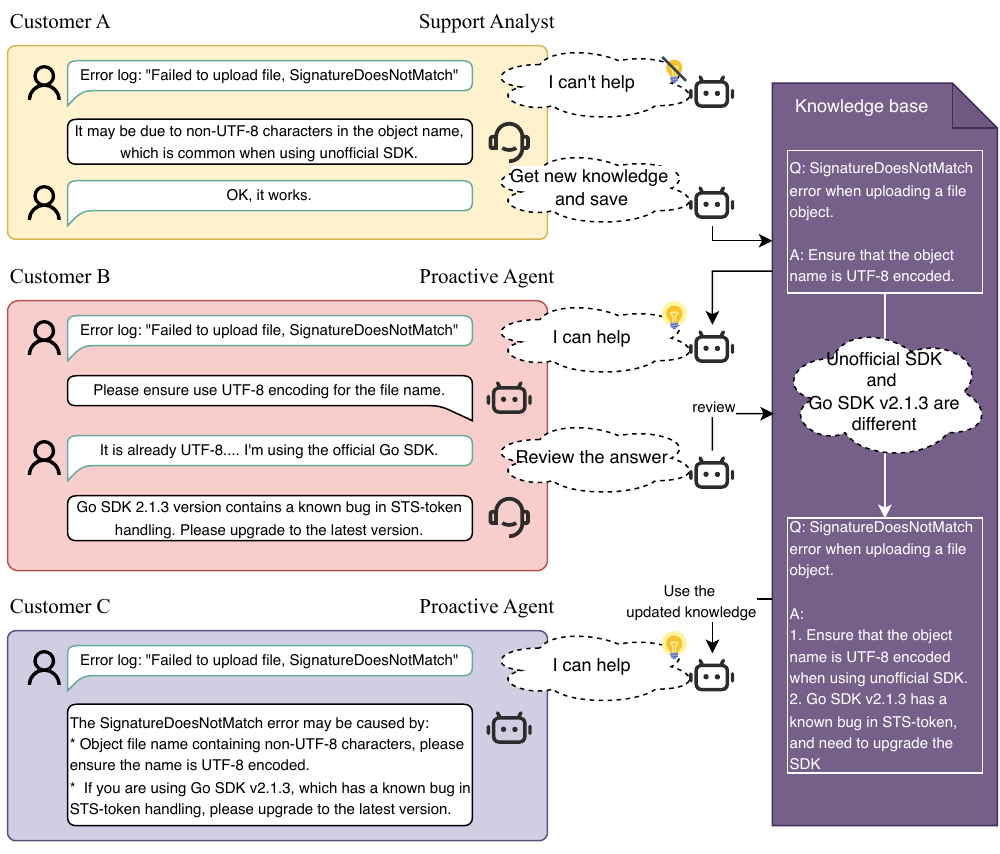}
    \caption{\name{} reviews the unaccepted answer and refines its knowledge base (simplified and anonymized).}
    \vspace{-2mm}
    \label{fig:case2}
\end{figure}

In real-world troubleshooting scenarios on Volcano Engine, each individual on-call dialogue is inherently constrained by its specific context.
These dialogues often begin with a description of the observed phenomenon, while the underlying root causes and solutions are then gradually uncovered through discussion.
The contextual details from such isolated cases are often insufficient to form a comprehensive knowledge-base entry.
Instead, a generalizable solution typically emerges only after synthesizing information from multiple related on-calls.

A case from July 2025 involving an object file upload error with a \textit{SignatureDoesNotMatch} code illustrates this process.
As shown in Figure \ref{fig:case2}, when this problem occurs, a customer initially only sees an error message and opens an on-call with this minimal log.
During the first occurrence, \name{} was unable to provide an immediate solution.
By following the subsequent dialogue, it learned that this was common in unofficial SDKs and could be resolved by encoding the file name in UTF-8 to match the signature.
When a second customer later encountered the same issue, \name{} proactively suggested the UTF-8 encoding solution.
Nevertheless, the answer was not accepted by the customer. 
Within our framework, such non-acceptance triggers a review mechanism, preventing potentially incorrect knowledge from being reused in future cases.

At the review stage in self-imporovement, \name{} analyzed the differences between these two on-calls involving the same error symptoms.
It identified that the second customer was using the Go SDK v2.1.3 with a known bug affecting the handling of STS tokens.
Thus, while the problem manifestation appeared identical, the root causes were different, necessitating a distinct solution.
The self-improvement module identified this discrepancy and executed an \textit{Update} operation, refining the existing knowledge entry with these new contextual constraints.

This case study demonstrates that identical symptoms in on-call support can stem from disparate root causes.
Such a self-improvement loop is essential for enabling \name{} to iteratively filter out mismatched knowledge and continuously enhance the accuracy and contextual relevance of its learned knowledge.

\section{Discussion and Lessons Learned}
\label{sec:discussion}
Our experience deploying \name{} at Volcano Engine yielded several generalizable lessons.

\textbf{The Criticality of Proactive Response Latency.}
A key lesson from our deployment is that proactive response speed is important.
The core value of \name{} is integrating into the live on-call dialogue between a customer and a human analyst.
To be effective, its answers must be delivered before the human analyst formulates a response.
This real-time requirement imposes a significant design constraint. It means that complex reasoning strategies such as multi-step planning, reflection, or chained tool usage, while powerful in other agent applications, are not suitable for on-call due to their inherent latency.
A more sophisticated answer that arrives too late is less valuable than a timely one.

\textbf{The Ambiguity of Explicit Feedback.} Explicit interactions, such as clicking \textit{Accept}, provide high-confidence positive signals.
However, an unaccepted answer is inherently ambiguous and does not necessarily indicate low answer quality. 
We observe that users often overlook feedback mechanisms due to operational factors, such as high-pressure troubleshooting environments or habitual focus on the chat interface rather than surrounding UI components.
Early versions of our work that relied solely on these explicit feedback as a universal metric for knowledge maintenance would introduce a significant conservative bias, leading the system to erroneously discard valid and helpful knowledge simply because it was not formally acknowledged.
To mitigate this issue, $\name{}$ treats explicit positive feedback as a sufficient—but not necessary—condition for validation.
For cases where feedback is absent, we decouple knowledge evolution from user actions by employing a dialogue-aware review mechanism. 
By analyzing the subsequent conversation context, the system can autonomously distinguish between truly incorrect answers and helpful but unacknowledged ones. 
This design ensures that the self-improvement loop remains robust and data-driven, rather than being constrained by the users' feedback behavior.

\textbf{The Evolving Impact of Foundation Models.}
Our experiments show that system performance is strongly influenced by the choice of foundation model. 
Accordingly, the current modular design of \name{} structures the LLM’s behavior by decomposing complex tasks into a sequence of simpler steps.
This approach is particularly beneficial for foundation models that struggle with complex, multi-part instructions, enabling substantial performance gains. 
However, as foundation models continue to improve in reasoning and instruction-following, the need for explicit architectural decomposition may diminish. 
In such cases, sufficiently capable models may reliably perform both identification and generation within a single, streamlined prompt, simplifying the overall system. 
For now, our results suggest that a modular approach remains the most robust and practical strategy for maximizing performance and ensuring reliability across the diverse landscape of currently available foundation models.

\section{Conclusion and Future Work}
\label{sec:conclusion}
In this paper, we identify key limitations of reactive agents: they are constrained by short action cycles and cannot learn effectively from ongoing, human-led incident resolutions. 
To address these limitations, we propose \name{}, a proactive agent–based system for on-call support deployed on the Volcano Engine cloud platform. 
\name{} enables proactive responses and supports continuous self-improvement throughout the entire on-call lifecycle. 
Our work establishes an efficient and practical paradigm for human–AI collaboration in on-call support.

Despite these advances, several challenges remain before proactive agents can fully close the loop in real-world on-call operations.
Future work will focus on extending proactive capabilities from providing answers to performing actions, such as automated diagnostics.
We will also investigate adaptive reasoning mechanisms that can balance the critical trade-off between low-latency responses and deep reasoning to further enhance the utility of \name{} in complex on-call scenarios.

\bibliographystyle{ACM-Reference-Format}
\bibliography{sample-base}

%
\appendix

\section{Prompt Template Details}

We use seed-1.6 model, a model with original thinking ability from ByteDance, to perform different tasks that introduced in this paper.
The model output constrains to a specific JSON schema, which aligns to the feature of structured output of seed-1.6 model.

\subsection{Prompt for Question Identification}
\label{appendix:question_identification}


\begin{lstlisting}
# Role
You are an intelligent on-call analysis expert of Volcano Engine.

# Task
Your task is to analyze the newly added messages from the customer, combing all the given messages and images in the on-call dialogue, and identify whether there is a question that within your ability scope.

# Your ability and responsibility scope
1. You are able to answer questions related to the Volcano Engine's product features, usage guidance, configuration instructions, and provide code examples.
2. You can help with explaining the error message, exception and common troubleshooting steps of Volcano Engine.

# Rules
1. If the messages contain a question that you are capable of answering, classify it as "Within Scope".
2. If the messages contain a question that is beyond your ability scope, classify it as "Out of Scope".
3. If the messages do not contain any question, classify it as "No assistance needed".
4. You just need to give the classification result, without answering the question.

Please analyze the newly added messages from the customer and give your classification result.
\end{lstlisting}

\subsection{Prompt for Answer Generation}
\label{appendix: answer_generation}

\begin{lstlisting}
# Role
You are an intelligent assistant. Please combine the historical dialogue with the references to understand and respond to the current question.

# References Usage Rules
Each reference is marked with <doc_n>xxx</doc_n>. Different reference materials are independent of each other.
You may only use a reference material if all the following conditions are satisfied:
* The target object is consistent.
* The issue phenomenon is consistent.
* The pre-conditions are consistent.

# Response Restrictions
* Do not ask the user for additional information.
* Do not include sensitive information (e.g., account, password, email, etc.).

# Response Guidelines
* Historical dialogue is only for understanding context; only answer the current question.
* When citing reference materials, embed the citation inline within the sentence using <doc_1><doc_2><doc_n> format.
* If unable to answer, reply with "Unable to answer" only, without additional explanations or content.
\end{lstlisting}

\subsection{Prompt for Answer Review}
\label{appendix:answer_review}

\begin{lstlisting}
# Role
You are an expert of reviewing knowledge base. You are reviewing the answers and the references that used for generating the answers.

# Task
Given a question(Q) from a on-call dialogue, answers for(A) this question that generated by a model, references that used for generated the answer from the knowledge base, and the follow-up dialogue of this question.
Your task is to review the answer and references, and select the corresponding action(Keep, Delete, Update) and operation tool to maintain the knowledge base according to the following rules.
Besides, if you need to update the references, you also need to provide the updated question and answer.

# Rules for the actions
1. Keep: If you find the answer from the follow-up dialogue is consistent with the existing answer, or the follow-up dialogue does not discuss about the question any more, select "Keep" which represents that you need to do nothing to the knowledge base.
2. Delete: If you find the answer from the follow-up dialogue has significantly differences from the existing answer, and the references is not suitable for this question, select "Delete" that you can delete the inappropriate references from the knowledge base.
3. Update: Historical references may contain some differences compared to the current on-call. If you find the answer from the follow-up dialogue has few differences from the existing answer, you need to distinguish the different background and prerequisites of this problem, and rewrite the question and answer to make them more accurate.
\end{lstlisting}

\end{document}